%% file: openbmb_main.tex
\documentclass[dvipsnames]{article}

\usepackage[final]{openbmb_2025}

\usepackage{amsmath}
\usepackage{amssymb}
\usepackage{mathtools}
\usepackage{amsthm}
\usepackage{multirow}
\usepackage{makecell}
\usepackage{subcaption}
\usepackage{wrapfig}
\usepackage{graphicx}
\usepackage{pifont}

\usepackage[utf8]{inputenc} 
\usepackage[T1]{fontenc}    
\usepackage{url}            
\usepackage{booktabs}       
\usepackage{amsfonts}       
\usepackage{nicefrac}       
\usepackage{microtype}      

\usepackage{color, soul}
\usepackage[most,breakable]{tcolorbox}
\usepackage[table]{xcolor}

\usepackage[labelfont=bf]{caption}
\usepackage{enumitem}
\usepackage{sectsty}

\usepackage{fancyhdr}
\renewcommand{\headrulewidth}{1pt}
\makeatletter
\def\headrule{{\if@fancyplain\let\headrulewidth\plainheadrulewidth\fi
\hrule\@height\headrulewidth\@width\textwidth \vskip-\headrulewidth}}
\makeatother

\definecolor{BMBDarkBlue}{HTML}{315EFE}
\definecolor{BMBLightBlue}{HTML}{00D3ED}
\usepackage[colorlinks, linkcolor=BMBDarkBlue, anchorcolor=BMBDarkBlue, citecolor=BMBDarkBlue, urlcolor=BMBDarkBlue]{hyperref}

\sectionfont{\color{BMBDarkBlue}\fontfamily{zi4}\selectfont}
\subsectionfont{\color{BMBDarkBlue}\fontfamily{zi4}\selectfont}
\subsubsectionfont{\color{BMBDarkBlue}\fontfamily{zi4}\selectfont}

\newtcolorbox{mytheorem}{
  colback=gray!5,       
  colframe=gray!80,     
  boxrule=0.5pt,        
  arc=4pt,              
  left=4pt,             
  right=4pt,            
  top=4pt,              
  bottom=4pt,           
}

\newcommand{\mymodel}[0]{MiniCPM-SALA}

\newcommand{\fancyheadname}{\textit{\textbf{\mymodel}}}

\title{\mymodel{}: Hybridizing Sparse and Linear Attention for Efficient Long-Context Modeling}

\author{%
MiniCPM Team
}

\def\huggingface{\raisebox{-1.5pt}{\includegraphics[height=1.05em]{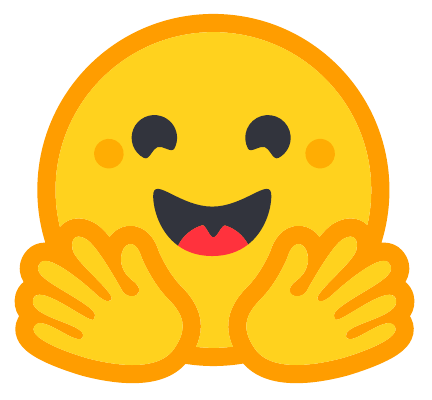}}}
\newcommand{\hflink}{https://huggingface.co/openbmb/\mymodel}
\def\github{\raisebox{-1.5pt}{\includegraphics[height=1.05em]{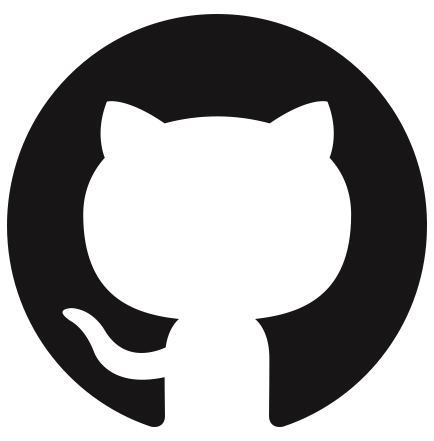}}}
\newcommand{\githublink}{https://github.com/OpenBMB/MiniCPM}

\begin{document}

\maketitle
\thispagestyle{fancy} 

\vspace{1em}
\input{sections/0-abstract}
\input{sections/1-intro}
\input{sections/2-methods}
\input{sections/3-exps}

\input{sections/4-conclusion}

\section{Contributions and Acknowledgments}
\mymodel{} is the result of the collective efforts of all members of our team. Please refer to \cite{halo} and \cite{infllmv2} for model architecture details.

\textbf{Contributors} (Ordered by the last name)\quad
Wenhao An, Yingfa Chen,
Yewei Fang, Jiayi Li, Xin Li, Yaohui Li, Yishan Li, Yuxuan Li, Biyuan Lin, Chuan Liu$^\star$, Hezi Liu, Siyuan Liu, Hongya Lyu,
Yinxu Pan, Shixin Ren, Xingyu Shen, Zhou Su, Haojun Sun, Yangang Sun, Zhen Leng Thai, Xin Tian,
Rui Wang$^\star$, Xiaorong Wang, Yudong Wang, Bo Wu, Xiaoyue Xu, Dong Xu, Shuaikang Xue, Jiawei Yang, 
Bowen Zhang, Jinqian Zhang, Letian Zhang, Shengnan Zhang, Xinyu Zhang, Xinyuan Zhang$^\star$, Zhu Zhang, Hengyu Zhao, Jiacheng Zhao$^\star$, Zhi Zheng, Jie Zhou, Zihan Zhou

\textbf{Project Design and Coordination}\quad
Shuo Wang, Chaojun Xiao, Xu Han, Zhiyuan Liu, Maosong Sun

\textbf{Affiliations}\quad
Contributors marked with $^\star$ are affiliated with XCORE SIGMA, while the remaining contributors are affiliated with OpenBMB.

\newpage

\bibliographystyle{citation}
\bibliography{citation}

\end{document}

%% file: sections/0-abstract.tex
\begin{abstract}
The evolution of large language models (LLMs) towards applications with ultra-long contexts faces challenges posed by the high computational and memory costs of the Transformer architecture. 
While existing sparse and linear attention mechanisms attempt to mitigate these issues, they typically involve a trade-off between memory efficiency and model performance.
This paper introduces \mymodel\footnote{SALA stands for Sparse Attention and Linear Attention.}, a 9B-parameter hybrid architecture that integrates the high-fidelity long-context modeling of sparse attention (InfLLM-V2) with the global efficiency of linear attention (Lightning Attention). 
By employing a layer selection algorithm to integrate these mechanisms in a 1:3 ratio and utilizing a hybrid positional encoding (HyPE), the model maintains efficiency and performance for long-context tasks.
Furthermore, we introduce a cost-effective continual training framework that transforms pre-trained Transformer-based models into hybrid models, which reduces training costs by approximately 75\% compared to training from scratch. Extensive experiments show that \mymodel\ maintains general capabilities comparable to full-attention models while offering improved efficiency. On a single NVIDIA A6000D GPU, the model achieves up to 3.5$\times$ the inference speed of the full-attention model at the sequence length of 256K tokens and supports context lengths of up to 1M tokens, 
a scale where traditional full-attention 8B models fail because of memory constraints.
\end{abstract}
\vspace{-0.5em}

%% file: sections/1-intro.tex
\section{Introduction}
\vspace{-1.0em}

As large language models (LLMs)~\citep{gpt4,gemini25,llama3,qwen3,deepseekv32} become increasingly effective, the application scenarios of LLMs are undergoing a profound paradigm shift, transitioning from simple question-answering~\citep{gpt3} to more advanced applications, such as deep understanding and generation of ultra-long contexts~\citep{longalign,longwriter,llmxmapreduce,storm}, repository-scale code engineering~\citep{deepseekcoder,swebench,repobench}, and long-horizon agents for complex tasks~\citep{chatdev,gaia,agencybench}.
For these advanced applications, models are no longer confined to processing fragmented information. Instead, they must demonstrate the capacity to handle ultra-long contexts, such as grasping entire technical manuals at once, analyzing comprehensive project dependency trees containing tens of thousands of lines of code, and maintaining coherent task states and memory over multi-day human-AI collaborations. This pursuit of holistic contextual information makes the ability to process millions of tokens a critical aspect for advanced LLMs~\citep{kimilinear,nemotron3nano}.

However, the Transformer architecture~\citep{Transformer}, which is the foundation of modern LLMs, encounters severe computational bottlenecks when handling ultra-long contexts due to its core full-attention mechanism. This bottleneck manifests primarily in two dimensions: (1) the {\em compute bottleneck} of computational complexity: for the standard attention mechanism, the computational cost grows quadratically with the sequence length $N$, i.e., its complexity is $\mathcal O(N^2)$. When the context scales to the level of millions of tokens, the huge overhead causes the inference latency to increase dramatically; (2) the {\em memory bottleneck} of KV-Cache: during the auto-regressive generation process, the model must store the key and value states (KVs) of all historical contextual tokens to avoid redundant computation. For a typical 8B-parameter model, even when utilizing Grouped Query Attention (GQA)~\citep{gqa}, the KV-Cache required for millions of tokens can reach dozens or even hundreds of gigabytes.

To address the aforementioned challenges, existing solutions have developed two primary paradigms: Sparse Attention~\citep{nsa,deepseekv32,infllm,infllmv2} and Linear Attention~\citep{gla,mamba,rwkv,deltanet,gateddeltanet}. Both paradigms present distinct advantages and inherent limitations. 
Sparse attention methods attempt to break the compute bottleneck by computing only the most salient portions of the attention matrix, such as adopting sliding windows or global anchors. However, these methods are hindered by a ``sparse computation, dense storage'' limitation. While local computation reduces immediate processing overhead, the model must still retain the full KV-Cache to support contextual information retrieval.
Linear attention utilizes recurrent formulations to successfully reduce computational complexity to $\mathcal O(N)$. 
Nevertheless, this extreme efficiency is achieved by the lossy compression of contextual information and inevitably results in performance degradation.

\mymodel\ employs a hybrid architecture of sparse and linear attention~\citep{halo}, specifically designed to achieve efficient ultra-long sequence modeling. This architecture combines the high-fidelity long-context modeling capabilities of InfLLM-V2~\citep{infllmv2} and the global computational efficiency of Lightning Attention~\citep{lightningattention}.
Through this integrated approach, the model significantly mitigates inference overhead and memory consumption, while simultaneously addressing the precision bottleneck typical of pure linear architectures in long-range information processing. 
Consequently, \mymodel\ provides a balanced solution that maintains both efficiency and high performance for long-context tasks. Furthermore, we employ the continual training paradigm to transform a pre-trained Transformer model into our hybrid model. By eschewing training from scratch, this approach significantly reduces the computational costs of model development.
While several works have begun exploring the integration of sparse and linear attention~\citep{hu2025hardwarealignedhierarchicalsparseattention, hou2025rwkvxlinearcomplexityhybrid, he2025alleviatingforgetfulnesslinearattention}, to the best of our knowledge, \mymodel\ is the first to demonstrate through large-scale experimentation that these hybrids can match the performance of full-attention baselines. Furthermore, the model exhibits high efficiency and strong performance in long-context processing.

In summary, the main contributions of this study can be outlined as follows:
\begin{itemize}[leftmargin=*]
    \item We introduce a Sparse-Linear hybrid attention mechanism integrating 25\% InfLLM-V2 and 75\% Lightning Attention to strike a balance between throughput and precision. By leveraging the granular focus of sparse attention for local details and the $\mathcal O(N)$ efficiency of linear attention for broad context, the architecture maintains high semantic accuracy as the sequence length scales up.
    \item We demonstrate that the Transformer-to-hybrid paradigm is a highly effective strategy for building strong hybrid models. This approach circumvents the inefficiencies of cold-start training by performing an architectural transformation on the pre-trained weights, thereby reducing the total training budget to approximately 25\% relative to training a comparable model from scratch.
    \item We adopt HyPE (Hybrid Positional Encoding)~\citep{halo} to effectively harmonize the performance across both short and long contexts. While maintaining general capabilities (e.g., knowledge, mathematics, and coding) comparable to modern full-attention models like Qwen3-8B, \mymodel\ has substantial advantages across multiple long-context benchmarks.
    \item \mymodel\ demonstrates substantial resource savings and speed advantages in long-context scenarios.
    On the NVIDIA A6000D GPU, \mymodel\ achieves up to 3.5$\times$ the inference speed of Qwen3-8B at a sequence length of 256K tokens. Furthermore, \mymodel\ supports inference at context lengths of up to 1M tokens on both NVIDIA A6000D and 5090 GPUs, whereas Qwen3-8B fails at this length due to out-of-memory (OOM) errors. These results demonstrate the broad prospects of \mymodel\ in edge-side information-intensive applications.
\end{itemize}

%% file: sections/2-methods.tex
\vspace{-1.0em}
\section{Model Development}
\vspace{-1.0em}

In this section, we introduce the model architecture and training strategies for \mymodel{}. Specifically, we combine the efficient sparse attention for long-context modeling and linear attention for global efficiency in \mymodel{}. Moreover, we also introduce an efficient training method, which can transform a standard Transformer model into sparse-linear hybrid attention.

\begin{figure*}[t]
    \centering
    \includegraphics[width=0.7\textwidth]{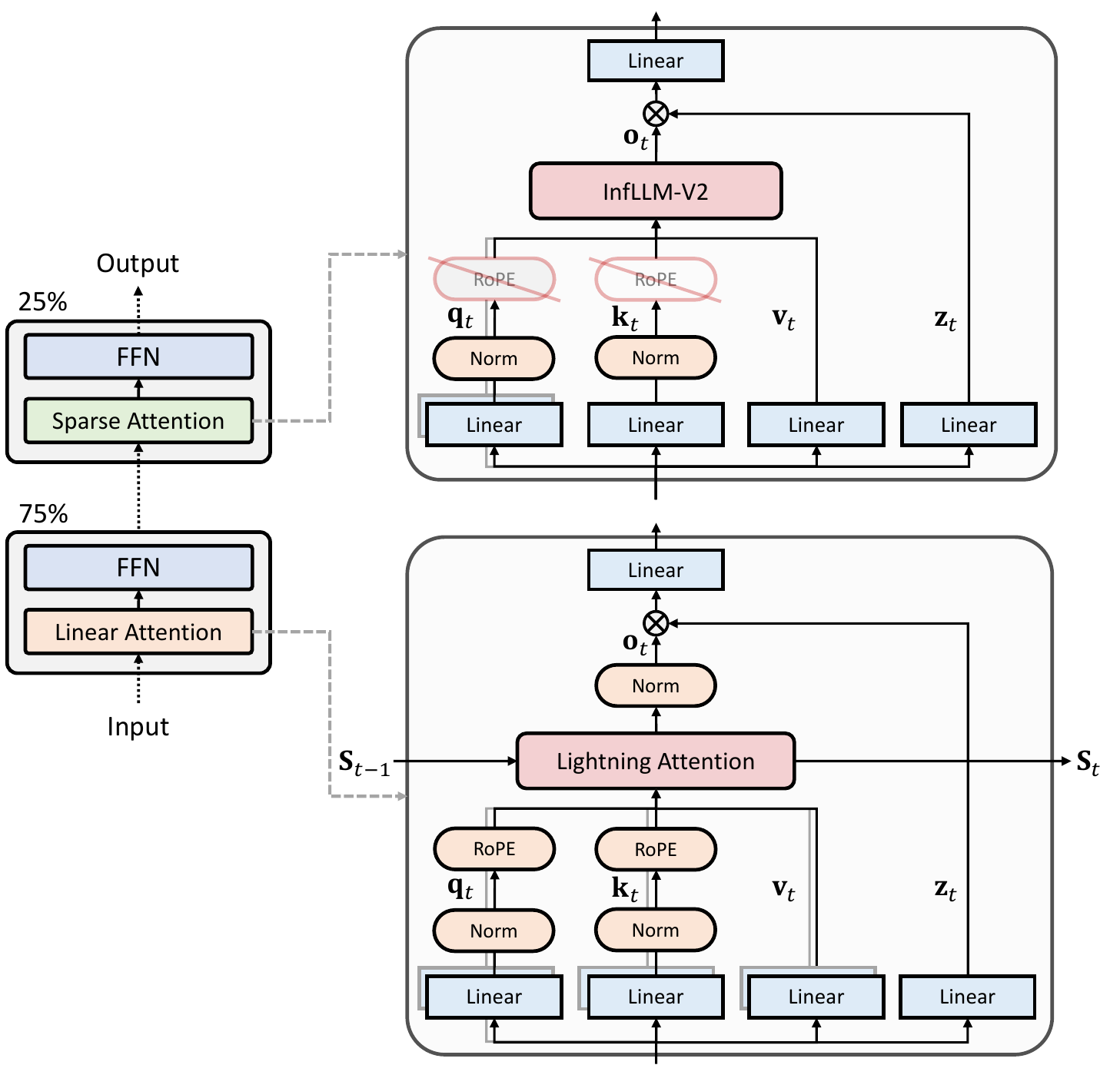}
    \caption{Architecture of \mymodel. The model adopts an efficient hybrid design that combines InfLLM-V2~\citep{infllmv2} and Lightning Attention~\citep{lightningattention} modules in a 1:3 ratio. Building on an intermediate MiniCPM-4.0~\citep{minicpm4} checkpoint, \mymodel\ undergoes a continual training phase to convert a standard Transformer model into a sparse-linear hybrid model.}
    \label{fig:arch}
\end{figure*}

\subsection{Model Architecture}
\label{subsec:model-arch}

The overall architecture of \mymodel\ is illustrated in Figure~\ref{fig:arch}.
\mymodel\ adopts a hybrid architecture that interleaves sparse attention layers and linear attention layers.
We retain the Feed-Forward Network~(FFN) block after each attention block in the Transformer architecture to ensure high-capacity knowledge representation. 
Inspired by the architectural designs of recent representative studies, such as Qwen3-Next~\citep{qwen3-next} and Kimi-Linear~\citep{kimilinear}, as well as our internal small-scale preliminary experiments, we employ a 1:3 mixing ratio: 25\% of the layers adopt sparse attention while the remaining 75\% employ linear attention.

This hybrid configuration leverages the complementary strengths of both attention mechanisms. Linear attention layers have constant computational and memory complexities with respect to sequence length, facilitating efficient processing of long contexts.
On the other hand, sparse attention layers facilitate effective modeling of long-range dependencies. Rather than naively uniformly interleaving the two attention variants, we determine the placement of sparse attention modules using the layer selection mechanism proposed by \citet{halo}, which results in superior downstream performance.

\textbf{Training Strategy}\quad
Existing paradigms for training hybrid models generally fall into two categories: (1)~training from scratch~\citep{falconh1, qwen3-next, kimilinear, nemotron3nano} and (2)~converting a pre-trained Transformer model into a hybrid model via cross-architecture distillation~\citep{mambainllama, rad, kl-guided-layer-selection, jetnemotron}. Although training from scratch offers simplicity and maximum architectural flexibility, continual-training conversion is a more resource-efficient alternative that leverages parameter inheritance from established pre-trained models. 
By recycling pre-trained weights and representations, the continual-training method significantly reduces the immense computational cost typically associated with \textit{de novo} training, achieving competitive performance with a fraction of the budget. 
Accordingly, \mymodel\ leverages a conversion-based framework that uses continual training to adapt a Transformer into an efficient hybrid version while preserving its core capabilities.

\textbf{Sparse Attention and Linear Attention}\quad
For the sparse attention layers, we incorporate InfLLM-V2~\citep{infllmv2}, which offers the distinct advantage of introducing no additional parameters to the architecture. Its inherent flexibility and ability to switch seamlessly between dense and sparse modes are highly compatible with our conversion process. 
This compatibility facilitates a stable training initialization by allowing sparse modules to inherit dense weights without architectural discrepancies, ensuring that the conversion to a hybrid structure does not compromise the model capacity.
For the linear attention layers, we utilize Lightning Attention~\citep{lightningattention}. Given our Transformer-to-hybrid conversion paradigm, Lightning Attention is selected for its functional proximity to the standard softmax attention. This structural alignment is intended to mitigate the complexities of parameter adaptation, thereby preserving pre-trained knowledge and ensuring robust downstream performance. Lightning Attention also provides better length generalization capabilities according to \citet{halo}, which may improve data efficiency during long-context continual-training.

\textbf{Other Architectural Improvements}\quad
Following HypeNet~\citep{halo}, we also introduce several architectural modifications to enhance the expressivity and training stability of \mymodel. These include QK-Normalization~\citep{qknorm}, HyPE~\citep{halo}, and the integration of output gates. 
\begin{itemize}[leftmargin=*]
    \item \textbf{QK-Normalization:} This is applied to all attention layers (both sparse and linear layers) to prevent the activation spikes that often occur in long-context training and further improve and boost the expressivity of linear attention modules.
    \item \textbf{HyPE (Hybrid Positional Encoding):} To balance rich positional awareness and long-range information retention, we employ a hybrid approach to positional encoding. We apply Rotary Positional Embedding (RoPE)~\citep{rope} to the linear attention layers to facilitate position-sensitive memory, allowing the model to preserve the relative order of tokens within the global context. On the other hand, we remove RoPE in the sparse attention layers. This strategic omission prevents the decay of long-distance information often associated with RoPE, thereby enabling more precise recall over extended contexts.
    \item \textbf{Output gates:} Furthermore, we incorporate an output gate after each attention block (both sparse and linear). This architectural choice aligns with recent advances in the gated attention mechanism~\citep{gatedattention}, in which the output gate has been shown to effectively mitigate issues such as attention sink.
    By regulating the information flow, the output gate prevents excessive focus on specific tokens and ensures a more flexible distribution of attention weights. Empirically, we observe that integrating output gates into both linear and sparse attention significantly improves model stability and performance.
\end{itemize}

\begin{table}[t]
    \centering
    \small
    \caption{Overview of the whole training process to build \mymodel.}
    \begin{tabular}{lcccr}
        \toprule
        \multirow{2}{*}{\bf Stage} & \bf Trainable & \bf Sparse & \bf Sequence & \multirow{2}{*}{\bf \# Tokens} \\
        & \bf Parameters & \bf Attention & \bf Length & \\
        \midrule
        Architecture Conversion (HALO) & 
        Linear Attention & Disabled & 0.5K & 1.3B \\
        \cmidrule(lr){1-1} \cmidrule(lr){2-5}
        Continual Stable-Training & All Parameters & Disabled & 4K & 314.6B \\
        \cmidrule(lr){1-1} \cmidrule(lr){2-5}
        Short-Decay Training & All Parameters & Disabled & 4K & 1006.6B \\
        \cmidrule(lr){1-1} \cmidrule(lr){2-5}
        \multirow{3}{*}{Long-Decay Training} & \multirow{3}{*}{All Parameters} & \multirow{3}{*}{Enabled} & 32K & 102.2B\\
        & & & 160K & 62.9B\\
        & & & 520K & 50.6B\\
        \cmidrule(lr){1-1} \cmidrule(lr){2-5}
        \multirow{2}{*}{Supervised Fine-Tuning} & \multirow{2}{*}{All Parameters} & \multirow{2}{*}{Enabled} & 64K & 204.5B\\
        & & & 140K & 213.3B \\
        \bottomrule
    \end{tabular}
    \label{tab:conversion}
\end{table}

\subsection{Model Training}
\label{subsec:model-training}

The training of \mymodel\ is conducted through a multi-stage process that starts from an intermediate checkpoint of MiniCPM-4.0~\citep{minicpm4}, which has already been trained on 7T tokens. This methodology represents an extended implementation of Hybrid Attention via Layer Optimization (HALO)~\citep{halo}. In the initial phase, we use the HALO framework to convert softmax attention to linear attention. This conversion serves as the starting point for subsequent pipeline stages, including continual pre-training and post-training. By leveraging this approach, the model can transition from a dense architecture to a hybrid structure while preserving the general capabilities acquired during the backbone's earlier training phases. 
The entire conversion process, consisting of five stages, is shown in Table~\ref{tab:conversion}.
It is worth noting that the Transformer-to-hybrid training of \mymodel\ consumes approximately 2T tokens. This corresponds to roughly 25\% of the data volume required to train MiniCPM-4.0 from scratch (8T tokens).

\textbf{Architecture Conversion (HALO)}\quad
The first stage uses HALO to convert the Transformer model from a full attention architecture to a hybrid architecture. During this phase, the training configuration of \mymodel\ differs from the standard HALO approach in two aspects.
First, regarding layer selection, we keep the first and last layers unconverted to improve training stability. For the remaining layers, we utilize the HALO selection algorithm to determine which layers are preserved as softmax attention layers. These preserved softmax attention layers are subsequently trained as sparse attention in later stages. The second difference from standard HALO is that we do not perform the final fine-tuning step of the original HALO process. Instead, we conduct more extensive continual pre-training and post-training, which comprise the subsequent stages of our methodology.
The training process at this stage is highly efficient, using only 1.3B tokens with a sequence length of 512 tokens. Furthermore, only the converted linear-attention layers are trainable during this stage, while all other parameters remain frozen.

\textbf{Continual Stable-Training}\quad
The second stage is continual stable-training. We use the checkpoint from the previous stage as the starting point for further training on the MiniCPM-4.0 pre-training dataset. The primary objective of this phase is to facilitate better coordination between the converted linear attention layers and other model components, including full attention layers, FFN layers, and embeddings.
The sequence length for this process is set to 4K tokens, with a total training volume of 314.6B tokens. Since the sequence length remains relatively short, the sparse attention is disabled at this stage to maintain computational efficiency. For the hyperparameter configuration, the learning rate~(LR) is set to $7.5 \times 10^{-3}$ and held constant after a 2,000-step LR warmup period. Accounting for the sequence length and the number of GPUs, the global batch size is set to 7.8M tokens.

\textbf{Short-Decay Training}\quad
The third stage is short-decay training, during which the LR undergoes exponential decay from $7.5 \times 10^{-3}$ to $3.75 \times 10^{-4}$. This process utilizes a sequence length of 4K tokens and a global batch size of 7.8M tokens. This stage involves training on 1T tokens, representing the most extensive data volume in the entire development pipeline. 
Building on the MiniCPM-4.0 decay strategy, we significantly increase the weight of L2 high-quality selection data~\citep{wang2026datasciencetechnologyagi} and introduce a large volume of PDF corpora and L3 synthetic data. 
This approach aims to enhance general capabilities and logical reasoning using high-information-density training data, achieving the efficient compression and internalization of massive amounts of knowledge.

\textbf{Long-Decay Training}\quad
The fourth stage, long-decay, progressively extends the context length from 4K to 32K, 160K, and finally 520K tokens. These processes use data volumes of 102.2B tokens, 62.9B tokens, and 50.6B tokens, respectively.
To accommodate the increased sequence lengths, the global batch size is adjusted to 7.8M, 9.8M, and 10.1M tokens, while the LR is systematically decays from $3 \times 10^{-4}$ to $2 \times 10^{-4}$ at 32K, then to $1 \times 10^{-4}$ at 160K, and finally to $3.75 \times 10^{-5}$ at 520K to conclude the process. 
At this stage, we up-sample the proportion of long-context data to better align the model with long-sequence distributions.
Given the growing computational advantages of sparse attention at longer sequences, we enable the sparse attention mechanism at this stage and maintain full-parameter training, thereby allowing the model to effectively learn the synergy between sparse attention and linear attention.

\textbf{Supervised Fine-Tuning}\quad
The SFT corpus for this stage is composed of high-quality reasoning-intensive data, encompassing code, mathematics, knowledge, function calls, and general dialogue. This selection is designed to fully catalyze the reasoning and task-execution capabilities under complex logic.
Furthermore, we specifically synthesize long-context data to enhance the precision of information retrieval and cross-document comprehension within extended sequences.
During the SFT stage, the context length is set to 64K and increased to 140K afterwards, utilizing 204.5B and 213.3B tokens, respectively. Sparse attention remains enabled throughout this entire process. By bridging shorter and longer contexts, this strategy allows the model to better balance general capabilities with long-context proficiency. For both phases, the LR follows a schedule with a 1,000-step warmup to a peak of $1 \times 10^{-3}$ before decaying to $1 \times 10^{-4}$, while the global batch sizes are set to 15.7M for the 64K phase and 17.8M for the 140K phase.

%% file: sections/3-exps.tex
\vspace{-1.0em}
\section{Experiments}
\vspace{-1.0em}

\begin{table}[t]
    \centering
    \small
    \caption{Standard evaluation results of \mymodel\ and other open-source LLMs.}
    \resizebox{0.999\textwidth}{!}{
    \begin{tabular}{l ccccc c}
        \toprule
        \bf Models & Qwen3 & Nemotron-Nano-v2 & MiniCPM-4.1 & Ministral-3-R & Falcon-H1R & \mymodel \\
        \cmidrule(lr){1-1} \cmidrule(lr){2-2} \cmidrule(lr){3-3} \cmidrule(lr){4-4} \cmidrule(lr){5-5} \cmidrule(lr){6-6} \cmidrule(lr){7-7} 
        \bf \# Param. & 8B & 9B & 8B & 8B & 7B & 9B \\
        \cmidrule(lr){1-7}
        \multicolumn{7}{c}{Knowledge} \\
        \cmidrule(lr){1-7}
CMMLU & 81.68  & 61.59  & 84.72  & 71.74  & 63.55  & 81.55 \\
MMLU-Pro & 73.26  & 71.79  & 72.70  & 68.75  & 70.98  & 67.04 \\
        \cmidrule(lr){1-7}
        \multicolumn{7}{c}{Code} \\
        \cmidrule(lr){1-7}
HumanEval & 93.90  & 93.90  & 91.46  & 96.95  & 96.34  & 95.12 \\
LCB-v5 & 56.89  & 68.26  & 56.89  & 65.87  & 67.66  & 60.48 \\
LCB-v6 & 48.57  & 60.00  & 51.43  & 53.71  & 57.71  & 52.00 \\
MBPP & 81.32  & 93.39  & 91.05  & 94.16  & 91.05  & 89.11 \\
        \cmidrule(lr){1-7}
        \multicolumn{7}{c}{Math} \\
        \cmidrule(lr){1-7}
AIME24 & 73.33  & 71.67  & 80.83  & 81.46  & 86.67  & 83.75 \\
AIME25 & 66.67  & 56.67  & 72.08  & 75.00  & 81.04  & 78.33 \\
        \cmidrule(lr){1-7}
        \multicolumn{7}{c}{Other} \\
        \cmidrule(lr){1-7}
BBH & 74.17  & 74.28  & 82.68  & 64.39  & 63.17  & 81.55 \\
IFEval & 84.66  & 86.69  & 77.45  & 70.06  & 86.32  & 76.34 \\
        \cmidrule(lr){1-7}
Average & 73.45  & 73.82  & 76.13  & 74.21  & 76.45  & \textbf{76.53} \\
        \bottomrule
    \end{tabular}}
    \label{tab:standard-result}
    \vspace{-1em}
\end{table}

\begin{table}[t]
    \centering
    \small
    \caption{Long-context evaluation results of \mymodel\ and other open-source LLMs.}
    \resizebox{0.999\textwidth}{!}{
    \begin{tabular}{lc rrrr r}
        \toprule
        \multicolumn{2}{c}{\bf Models} & Qwen3 & Nemotron-Nano-v2 & Ministral-3-R & Falcon-H1R & \mymodel \\
        \cmidrule(lr){1-2} \cmidrule(lr){3-3} \cmidrule(lr){4-4} \cmidrule(lr){5-5} \cmidrule(lr){6-6} \cmidrule(lr){7-7}
        \multicolumn{2}{c}{\bf \# Param.} & 8B & 9B & 8B & 7B & 9B \\
        \cmidrule(lr){1-7}
\multirow{2}{*}{RULER} & 64K & 80.53  & 88.77  & 70.66  & 56.50  & 92.65 \\
& 128K & 71.74  & 68.01 & 45.09  & 36.33  & 89.37 \\
\cmidrule(lr){1-2} \cmidrule(lr){3-6} \cmidrule(lr){7-7}
\multirow{6}{*}{MRCR} & 64K-2N & 29.20  & 20.91  & 44.02  & 13.18  & 29.77 \\
 & 64K-4N & 21.56  & 13.69 & 35.80  & 9.06  & 20.57 \\
 & 64K-8N & 17.82  & 13.24  & 17.23  & 6.93  & 16.56 \\
 & 128K-2N & 26.50  & 14.61  & 50.30  & 9.17  & 28.62 \\
 & 128K-4N & 14.75  & 12.20  & 22.66  & 8.22  & 19.62 \\
 & 128K-8N & 12.15  & 7.55  & 14.47  & 7.54  & 10.12 \\
 \cmidrule(lr){1-2} \cmidrule(lr){3-6} \cmidrule(lr){7-7}
\multirow{3}{*}{NoLiMa} & 32K & 43.40  & 19.69 & 3.78  & 14.89  & 54.54 \\
 & 64K & 23.35 & 11.82  & 2.48  & 9.87  & 42.95 \\
 & 128K & 11.25 & 5.80  & 3.48  & 4.73  & 23.86 \\
 \cmidrule(lr){1-2} \cmidrule(lr){3-6} \cmidrule(lr){7-7}
\multicolumn{2}{l}{Average} & 32.02  & 25.12 & 28.18  & 16.04  & \bf 38.97 \\
        \bottomrule
    \end{tabular}}
    \label{tab:long-result}
\end{table}

\begin{table}[ht]
    \centering
    \small
    \caption{Ultra-long context evaluation results of \mymodel\ and other open-source LLMs. $^*$ denotes results cited from the official Qwen3-Next documentation.}
    \begin{tabular}{l c c c c}
        \toprule
        \bf RULER & 128K & 512K & 1000K & 2048K \\
        \cmidrule(lr){1-1} \cmidrule(lr){2-5}
        Qwen3-30B-A3B-Instruct-2507$^*$ & 89.1 & 78.4 & 72.8 & - \\
        Qwen3-235B-A22B-Instruct-2507$^*$ & 93.9 & 90.9 & 84.5 & - \\
        Qwen3-Next-80B-A3B-Instruct$^*$ & 96.0 & 86.9 & 80.3 & - \\
        \cmidrule(lr){1-1} \cmidrule(lr){2-5}
        \mymodel\ (9B) & 89.4 & 87.1 & 86.3 & 81.6 \\
        \bottomrule
    \end{tabular}
    \label{tab:ultra-long-result}
\end{table}


\begin{figure}[ht]
    \centering
    \begin{subfigure}{0.48\textwidth}
        \includegraphics[width=\linewidth]{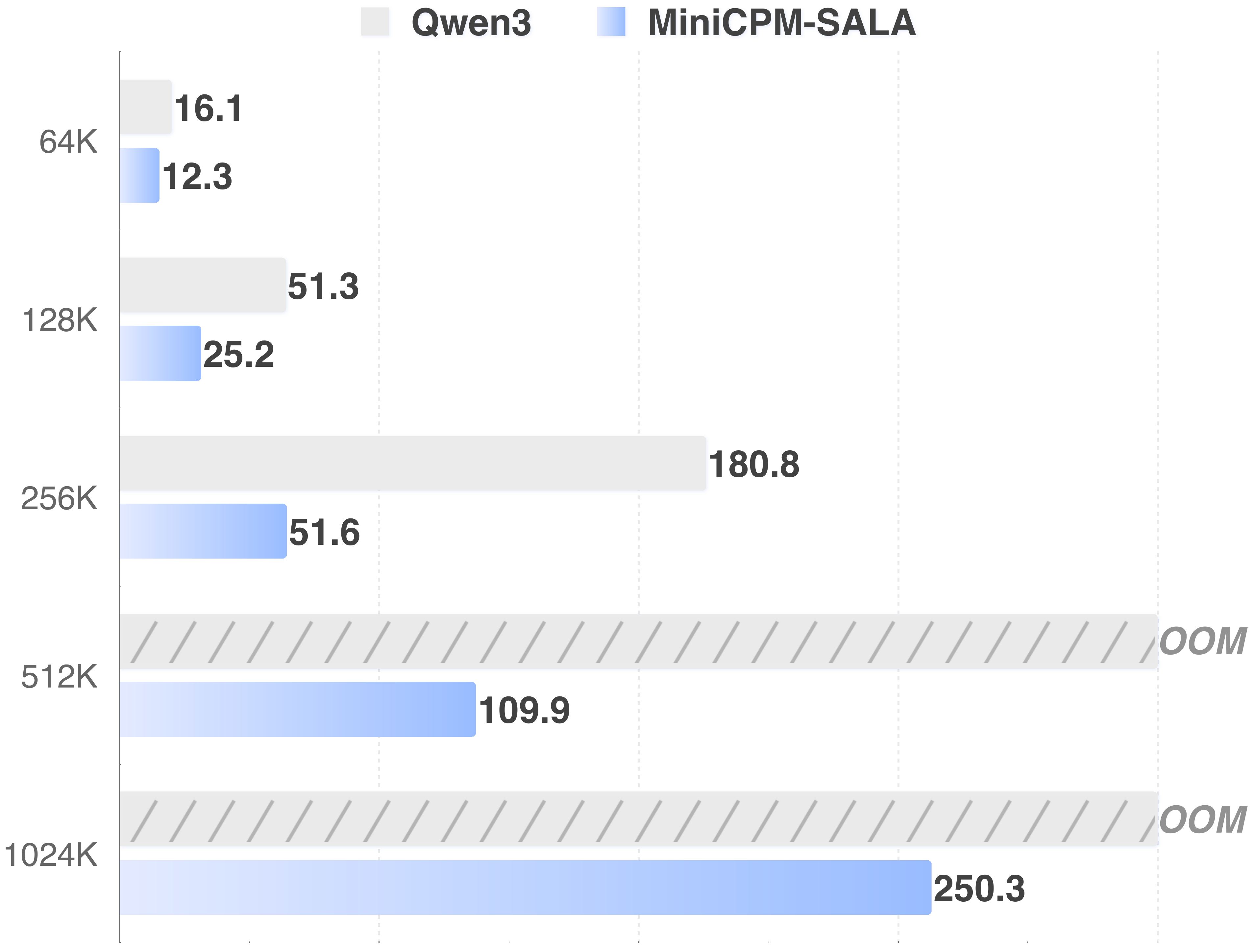}
        \captionsetup{justification=centerlast, singlelinecheck=false, width=\linewidth}
        \caption{TTFT (s) on A6000D (non-quantized).}
    \end{subfigure}
    \hfill
    \begin{subfigure}{0.48\textwidth}
        \includegraphics[width=\linewidth]{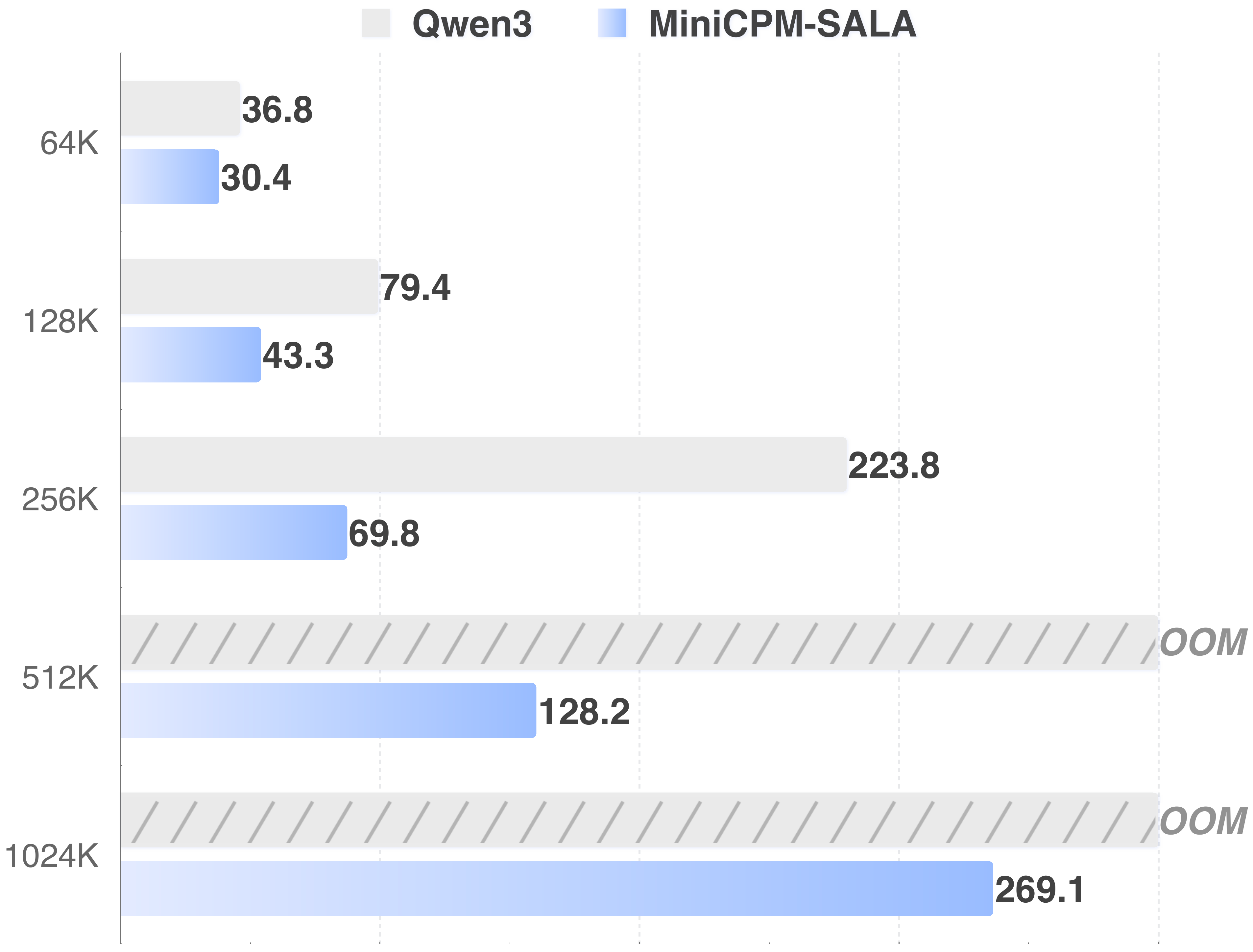}
        \captionsetup{justification=centerlast, singlelinecheck=false, width=\linewidth}
        \caption{End-to-end (s) latency on A6000D (non-quantized).}
    \end{subfigure}
    
    \vspace{0.5cm}
    
    \begin{subfigure}{0.48\textwidth}
        \includegraphics[width=\linewidth]{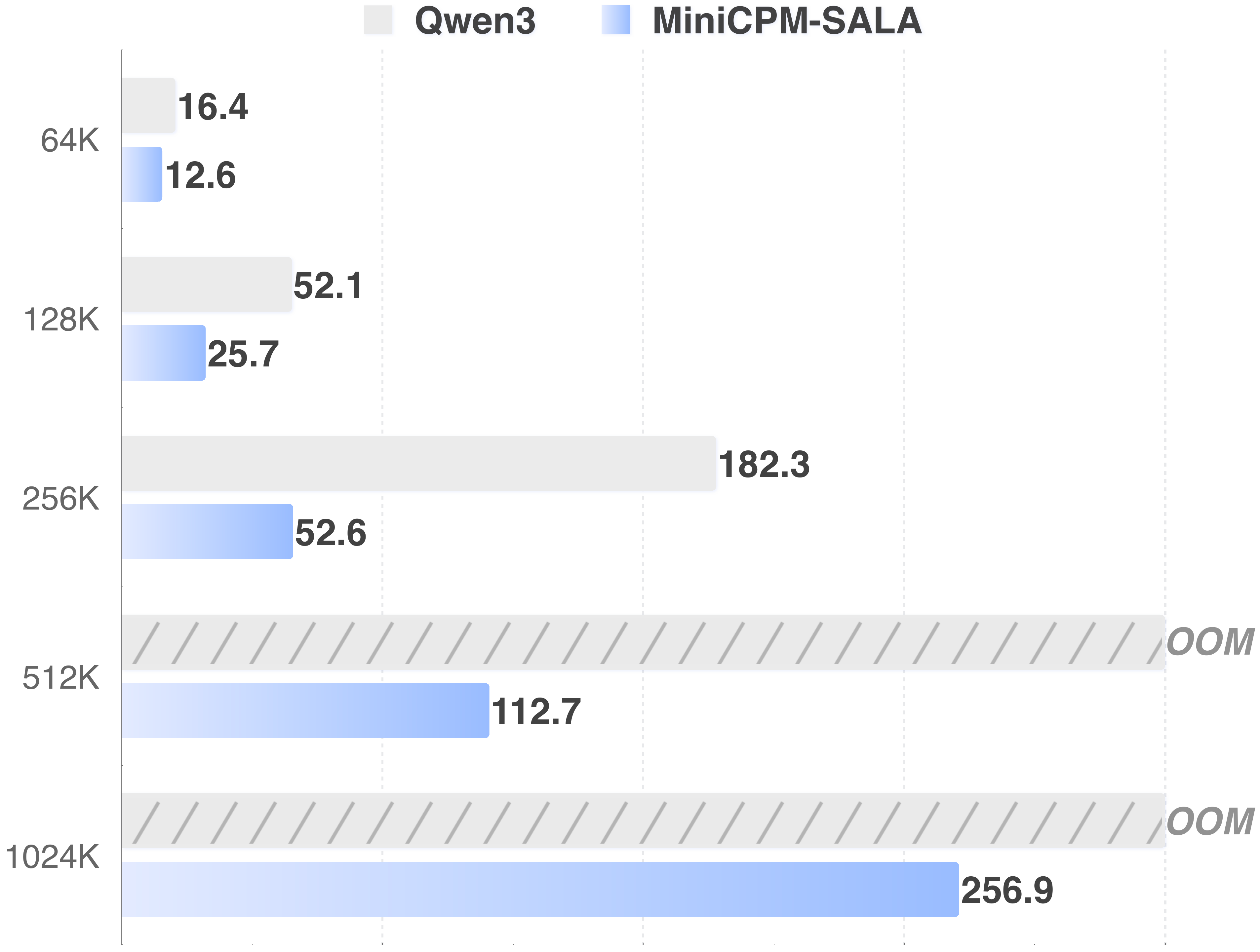}
        \captionsetup{justification=centerlast, singlelinecheck=false, width=\linewidth}
        \caption{TTFT (s) on A6000D (quantized).}
    \end{subfigure}
    \hfill
    \begin{subfigure}{0.48\textwidth}
        \includegraphics[width=\linewidth]{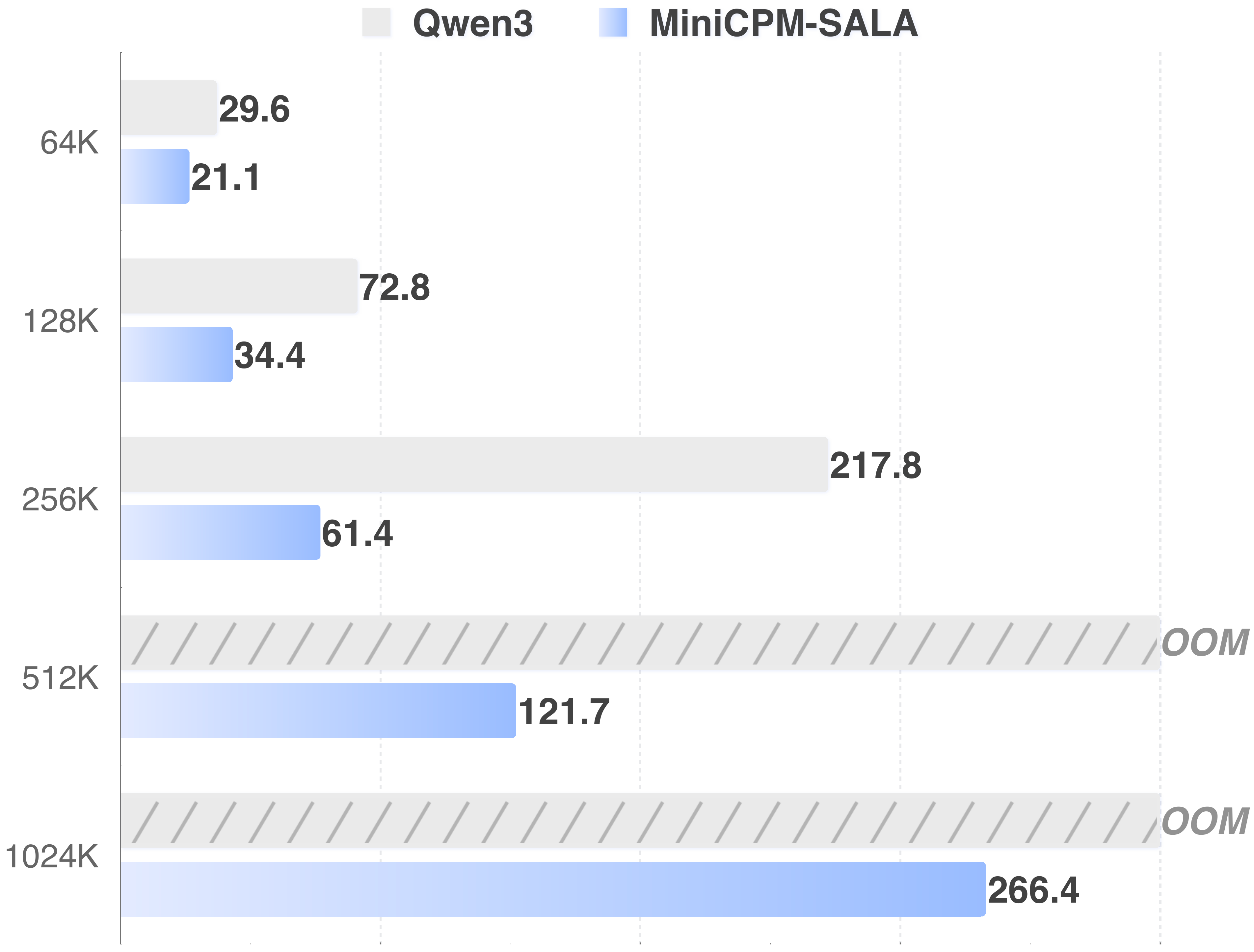}
        \captionsetup{justification=centerlast, singlelinecheck=false, width=\linewidth}
        \caption{End-to-end (s) latency on A6000D (quantized).}
    \end{subfigure}
    
    \caption{Inference speed comparison between Qwen3-8B and \mymodel. For each tested sequence length, the models process a specified input (prefilling) and generate 1K tokens (decoding). ``TTFT'' denotes Time To First Token, representing the prefilling latency, while ``End-to-end'' measures the total latency including both prefilling and decoding phases.}
    \label{fig:inference_speed_a6000d}
\end{figure}

\begin{figure}[ht]
    \centering
    \begin{subfigure}{0.48\textwidth}
        \includegraphics[width=\linewidth]{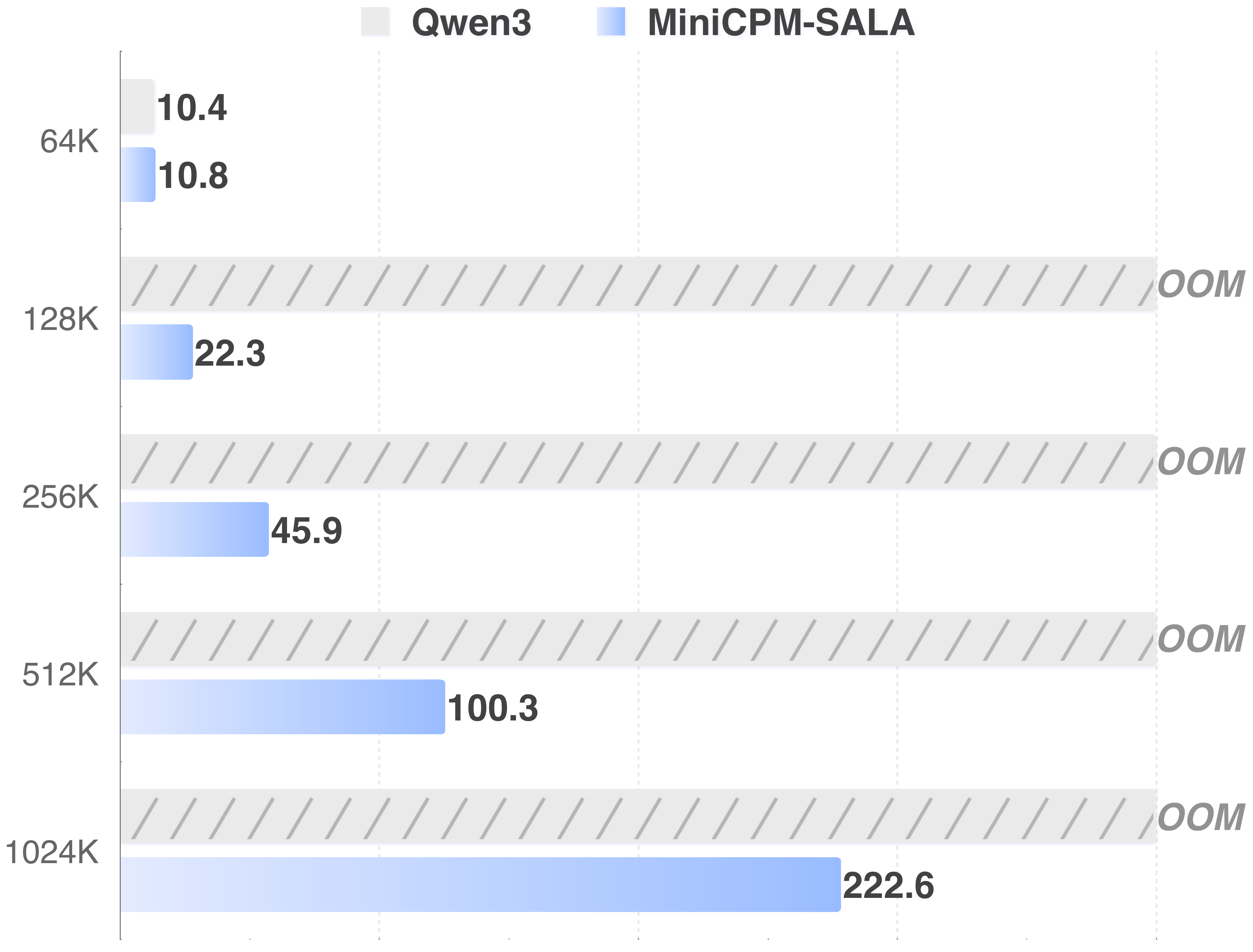}
        \captionsetup{justification=centerlast, singlelinecheck=false, width=\linewidth}
        \caption{TTFT (s) on 5090 (non-quantized).}
    \end{subfigure}
    \hfill
    \begin{subfigure}{0.48\textwidth}
        \includegraphics[width=\linewidth]{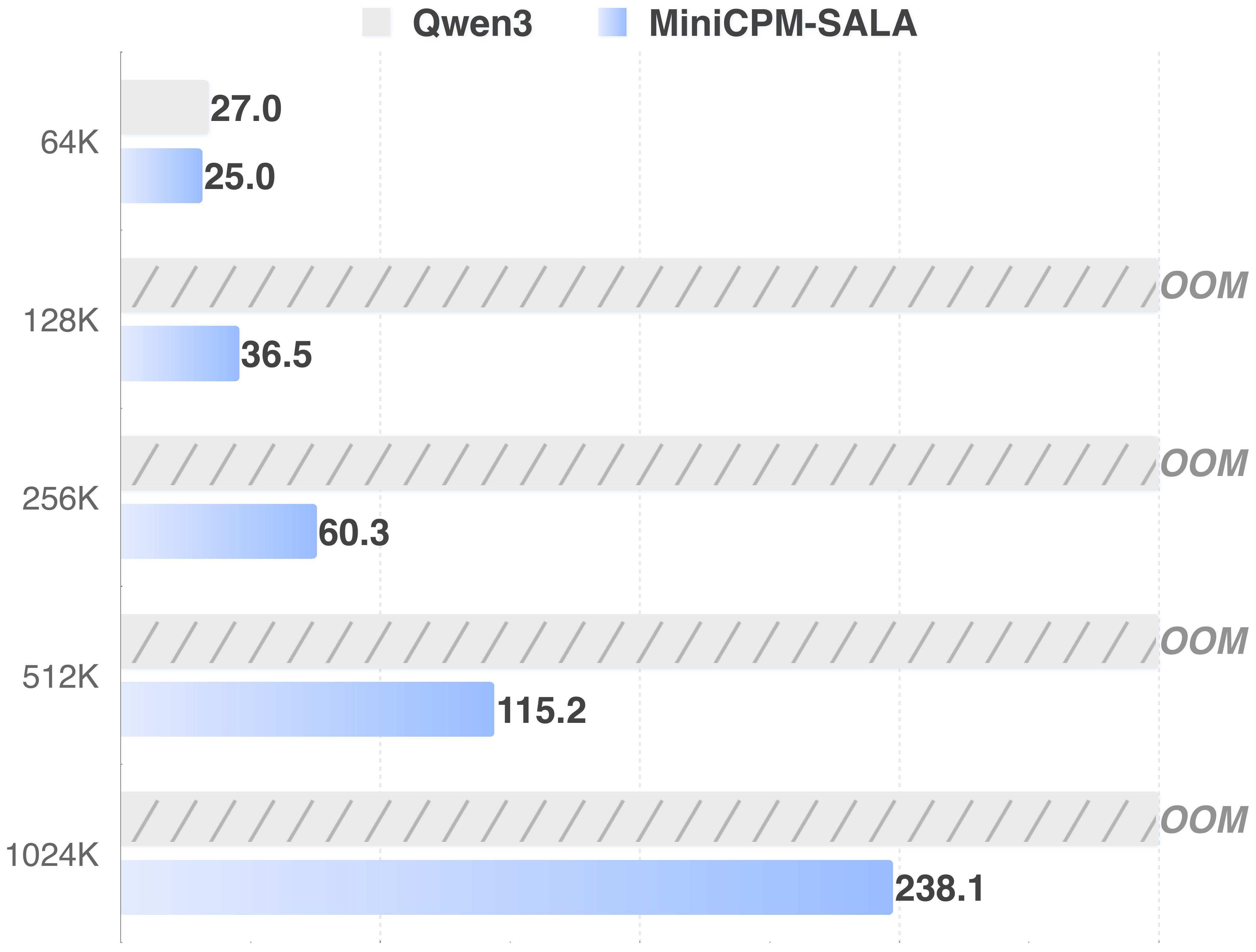}
        \captionsetup{justification=centerlast, singlelinecheck=false, width=\linewidth}
        \caption{End-to-end (s) latency on 5090 (non-quantized).}
    \end{subfigure}
    
    \vspace{0.5cm}
    
    \begin{subfigure}{0.48\textwidth}
        \includegraphics[width=\linewidth]{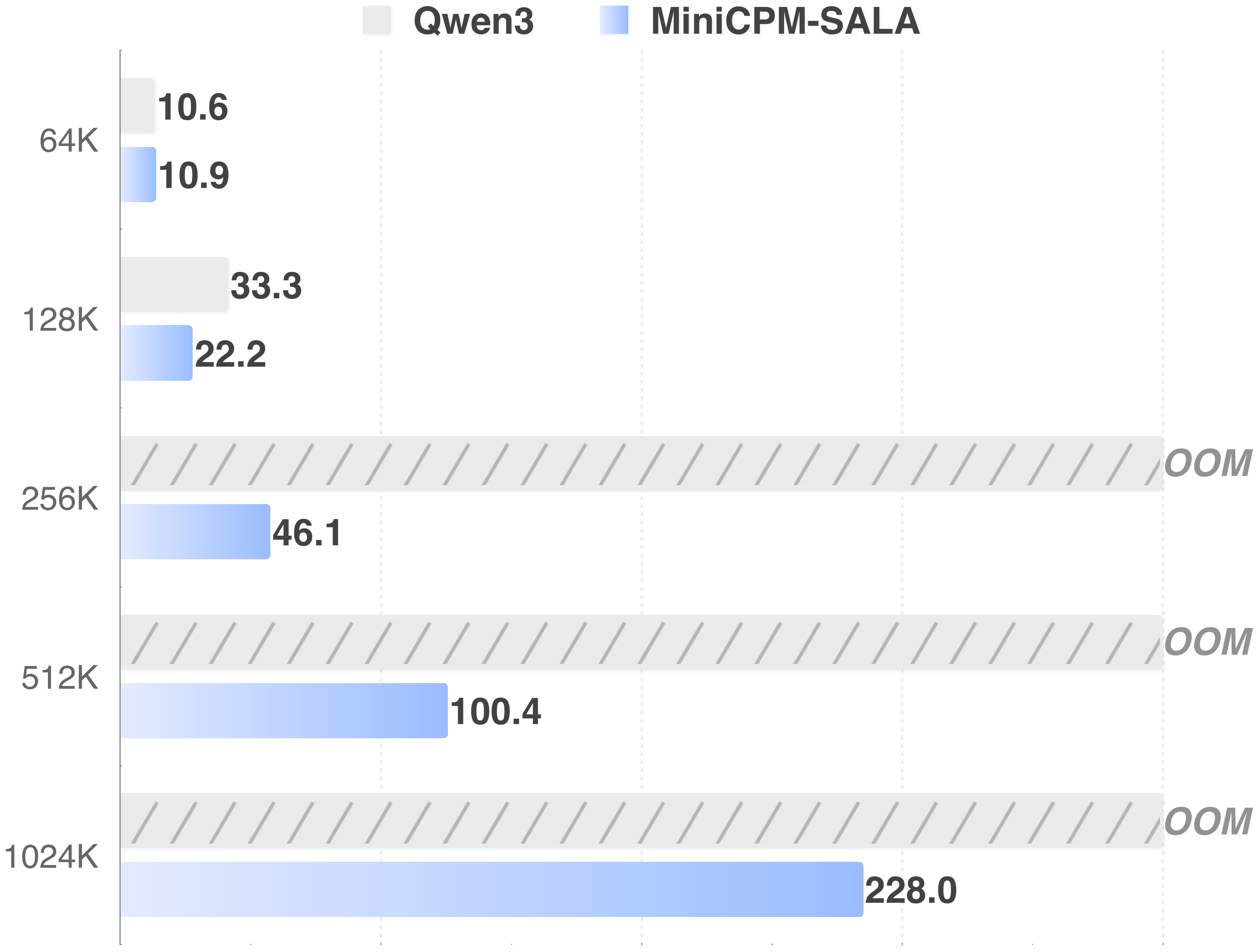}
        \captionsetup{justification=centerlast, singlelinecheck=false, width=\linewidth}
        \caption{TTFT (s) on 5090 (quantized).}
    \end{subfigure}
    \hfill
    \begin{subfigure}{0.48\textwidth}
        \includegraphics[width=\linewidth]{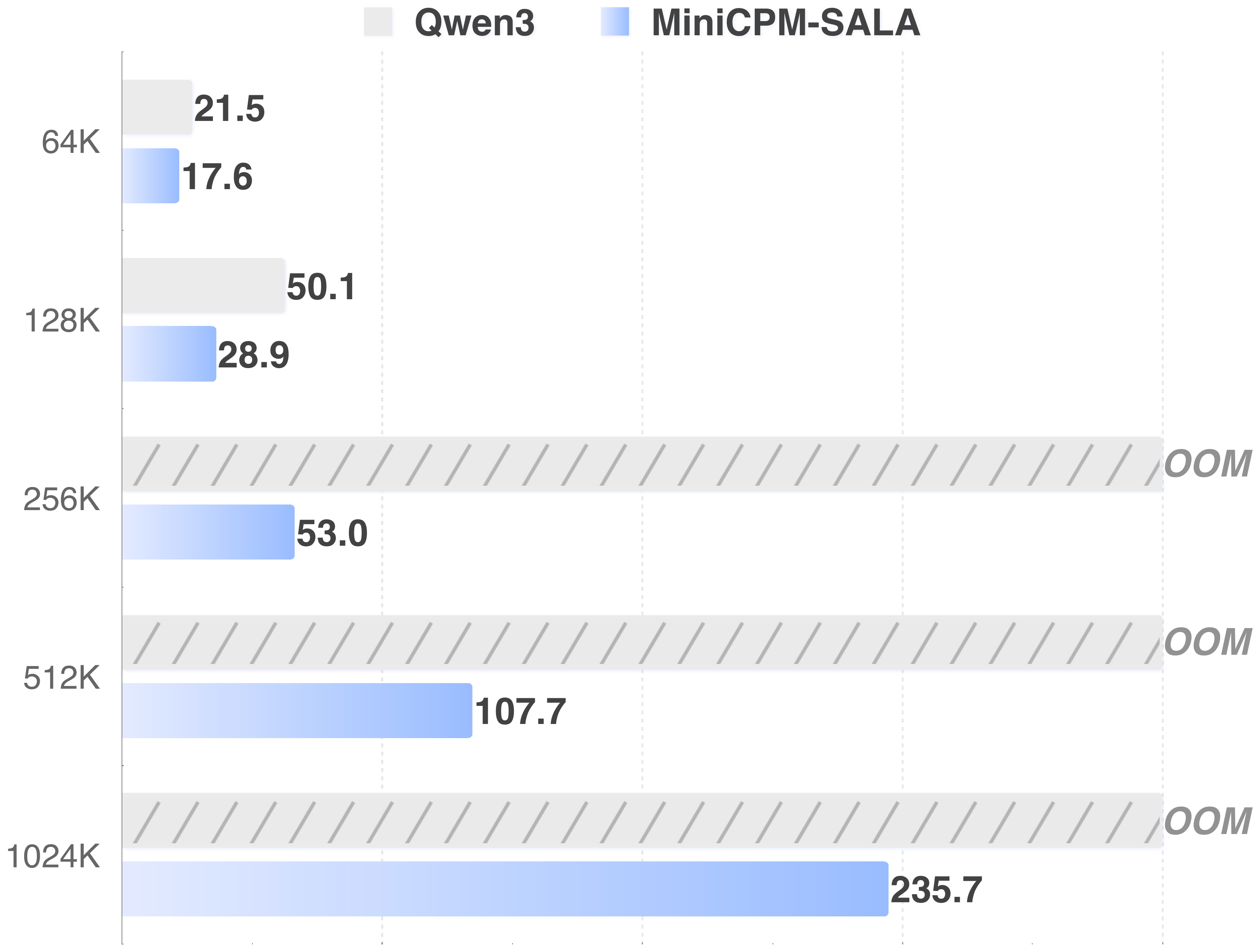}
        \captionsetup{justification=centerlast, singlelinecheck=false, width=\linewidth}
        \caption{End-to-end (s) latency on 5090 (quantized).}
    \end{subfigure}
    
    \caption{Inference speed comparison between Qwen3-8B and \mymodel. For each tested sequence length, the models process a specified input (prefilling) and generate 1K tokens (decoding).}
    \label{fig:inference_speed_5090}
\end{figure}

\subsection{Model Performance}

\paragraph{Benchmarks} To thoroughly assess the general capabilities of the model, we conducted evaluations across a diverse array of benchmarks. These include knowledge-intensive tasks (CMMLU~\citep{cmmlu}, MMLU-Pro~\citep{mmlupro}), coding benchmarks (HumanEval~\citep{humaneval}, LCB-v5/v6~\cite{lcb}, MBPP~\citep{mbpp}), and mathematical reasoning sets (AIME24/25~\citep{aime}), alongside other representative benchmarks such as BBH~\citep{bbh} and IFEval~\citep{ifeval}.
We further evaluated long-context capabilities using RULER~\citep{ruler}, MRCR\footnote{\url{https://huggingface.co/datasets/openai/mrcr}}, and NoliMa~\citep{nolima}. We utilized the OpenCompass framework~\citep{opencompass} to conduct the evaluations.

\textbf{Baseline Models}\quad
Given that \mymodel is a 9B-parameter model, we selected a series of modern baselines of comparable size, encompassing both hybrid and full-attention architectures. Specifically, the baselines include Qwen3-8B~\citep{qwen3}, Nemotron-Nano-v2-9B~\citep{nemotronnano2}, MiniCPM-4.1-8B~\citep{minicpm4}, Ministral-3-Reasoning-8B~\citep{ministral3}, and Falcon-H1R-7B~\citep{falconh1r}. We exclude MiniCPM-4.1-8B from the evaluation of long contexts because of its limitation to a context length of 64K.

\textbf{Results of Standard Evaluation}\quad
Table~\ref{tab:standard-result} presents the performance of \mymodel\ across a variety of standard benchmarks. The model achieves an average score of 76.53, which represents a competitive level among open-source models of a similar scale. In coding tasks, the model demonstrates high proficiency with scores of 95.12 on HumanEval and 89.11 on MBPP. Mathematical reasoning capabilities also remain robust, as evidenced by the scores of 83.75 on AIME24 and 78.33 on AIME25. These results indicate that the integration of long-context mechanisms does not result in a significant degradation of general capabilities or short-context performance. The model maintains a performance profile that is comparable to, and in some cases exceeds, the performance of models such as Qwen3-8B and Falcon-H1R-7B in standard evaluation settings.

\textbf{Results of Long-Context Evaluation}\quad
The evaluation of long-context capabilities is summarized in Table~\ref{tab:long-result}, covering benchmarks such as RULER, MRCR, and NoLiMa. \mymodel\ shows a notable proficiency in managing extended input sequences. On the RULER benchmark at a 128K context length, the model maintains a score of 89.37, while many other baselines exhibit a more pronounced decrease in accuracy at the same scale. The advantage of the model is particularly visible in the NoLiMa benchmark, where it achieves a score of 23.86 at the 128K level. This performance is substantially higher than the scores recorded for other models in the comparison. With an overall average long-context score of 38.97, the model demonstrates improved stability and effective information retrieval across large context windows.

\textbf{Results of Ultra-Long Context}\quad
As demonstrated in Table~\ref{tab:ultra-long-result}, \mymodel\ exhibits surprising length extrapolation capabilities. The results for the Qwen3 models are sourced from the official Qwen3-Next documentation\footnote{\url{https://huggingface.co/Qwen/Qwen3-Next-80B-A3B-Instruct}}.
Despite being restricted to a 520K training length, the model successfully extrapolates to 2048K tokens without a significant degradation in performance, maintaining a score of 81.6. It is worth noting that this extrapolation requires no auxiliary techniques (e.g., YaRN~\citep{yarn}). This result highlights the efficacy of our approach in handling context windows far beyond the training stage.
Additionally, \mymodel\ shows remarkable parameter efficiency, surpassing the performance of the Qwen3-Next-80B-A3B-Instruct model at the 1000K context length (86.3 vs. 80.3), proving that effective long-context processing does not necessarily require massive parameter counts.
The length extrapolation capabilities of \mymodel\ can be attributed to the NoPE configuration within the sparse attention layers. In this design, the stored KV-Cache does not require combination with positional information, which can otherwise hinder the capture of long-range dependencies.

\subsection{Inference Speed}

We assessed the inference speed of \mymodel\ and Qwen3-8B across different hardware and sequence lengths. To verify the long-text processing capabilities of the model in edge computing scenarios, we conducted experiments not only on cloud-grade inference chips, such as the NVIDIA A6000D, but also on consumer-grade edge GPUs, such as the NVIDIA 5090. For each sequence length, we measured both the Time To First Token (TTFT) and the end-to-end latency. The former serves as an indicator of the prefilling speed, while the latter reflects the combined performance of the prefilling and decoding phases. To align the evaluation with practical deployment scenarios, we assessed the inference latency for both non-quantized models and models compressed via GPTQ~\citep{gptq} INT4 quantization.

Figure~\ref{fig:inference_speed_a6000d} presents a comprehensive comparison of inference latency between Qwen3-8B and \mymodel\ on an NVIDIA A6000D GPU (96GB VRAM). We evaluated performance across sequence lengths ranging from 64K to 1024K tokens. As illustrated, \mymodel\ demonstrates a significant performance advantage over the baseline across all tested configurations. In non-quantized settings, \mymodel\ consistently achieves lower latency. Notably, at a sequence length of 256K, \mymodel\ reduces the TTFT from 180.8s (Qwen3) to just 51.6s.

Crucially, the results highlight a distinct advantage in memory efficiency. While Qwen3-8B encounters OOM failures at sequence lengths of 512K and 1024K, \mymodel\ successfully processes these extended contexts. For example, at 1024K tokens, \mymodel\ maintains a TTFT of 250.3s (non-quantized) and 256.9s (quantized), whereas the baseline fails to complete the inference. This trend persists in the end-to-end latency metrics, proving that \mymodel\ is robust enough for ultra-long context generation tasks where full-attention models fail.

Figure~\ref{fig:inference_speed_5090} demonstrates the critical advantage of \mymodel\ on memory-constrained hardware. On the RTX 5090 (32GB VRAM), the baseline Qwen3-8B hits a ``memory wall'' significantly earlier than on the A6000D, triggering OOM errors at just 128K tokens in non-quantized settings and 256K in quantized settings. In stark contrast, \mymodel\ successfully scales to 1024K context lengths without memory failure. This suggests that \mymodel\ effectively democratizes long-context inference, enabling 1M-token processing on consumer-level GPUs where full-attention architectures are unusable.

%% file: sections/4-conclusion.tex
\section{Conclusion}

In this paper, we presented \mymodel{}, a hybrid architecture that combines sparse and linear attention to overcome the computational and memory bottlenecks of ultra-long context modeling. By utilizing a cost-effective Transformer-to-hybrid training paradigm, we successfully retained the general capabilities of full-attention models while reducing training costs by approximately 75\%. Experimental results confirm that \mymodel{} achieves a substantial inference speedup and enables 1M-token context processing on single GPUs (e.g., NVIDIA A6000D), surpassing the limitations of standard 8B models. These results establish \mymodel{} as a scalable and accessible solution for next-generation, information-intensive applications.

%% file: citation.bib
@inproceedings{yarn,
title={Ya{RN}: Efficient Context Window Extension of Large Language Models},
author={Bowen Peng and Jeffrey Quesnelle and Honglu Fan and Enrico Shippole},
booktitle={The Twelfth International Conference on Learning Representations},
year={2024},
url={https://openreview.net/forum?id=wHBfxhZu1u}
}

@misc{he2025alleviatingforgetfulnesslinearattention,
      title={Alleviating Forgetfulness of Linear Attention by Hybrid Sparse Attention and Contextualized Learnable Token Eviction}, 
      author={Mutian He and Philip N. Garner},
      year={2025},
      eprint={2510.20787},
      archivePrefix={arXiv},
      primaryClass={cs.CL},
      url={https://arxiv.org/abs/2510.20787}, 
}

@misc{hu2025hardwarealignedhierarchicalsparseattention,
      title={Hardware-aligned Hierarchical Sparse Attention for Efficient Long-term Memory Access}, 
      author={Xiang Hu and Jiaqi Leng and Jun Zhao and Kewei Tu and Wei Wu},
      year={2025},
      eprint={2504.16795},
      archivePrefix={arXiv},
      primaryClass={cs.CL},
      url={https://arxiv.org/abs/2504.16795}, 
}

@misc{hou2025rwkvxlinearcomplexityhybrid,
      title={RWKV-X: A Linear Complexity Hybrid Language Model}, 
      author={Haowen Hou and Zhiyi Huang and Kaifeng Tan and Rongchang Lu and Fei Richard Yu},
      year={2025},
      eprint={2504.21463},
      archivePrefix={arXiv},
      primaryClass={cs.CL},
      url={https://arxiv.org/abs/2504.21463}, 
}

@misc{aime,
      title={{AIME} Problems and Solutions},
      author={{AIME}},
      year={2025},
      url={https://artofproblemsolving.com/wiki/index.php/AIME_Problems_and_Solutions}
}

@misc{ifeval,
      title={Instruction-Following Evaluation for Large Language Models}, 
      author={Jeffrey Zhou and Tianjian Lu and Swaroop Mishra and Siddhartha Brahma and Sujoy Basu and Yi Luan and Denny Zhou and Le Hou},
      year={2023},
      eprint={2311.07911},
      archivePrefix={arXiv},
      primaryClass={cs.CL},
      url={https://arxiv.org/abs/2311.07911}, 
}

@inproceedings{nolima,
title={NoLiMa: Long-Context Evaluation Beyond Literal Matching},
author={Ali Modarressi and Hanieh Deilamsalehy and Franck Dernoncourt and Trung Bui and Ryan A. Rossi and Seunghyun Yoon and Hinrich Schuetze},
booktitle={Forty-second International Conference on Machine Learning},
year={2025},
url={https://openreview.net/forum?id=0OshX1hiSa}
}

@inproceedings{ruler,
title={{RULER}: What{\textquoteright}s the Real Context Size of Your Long-Context Language Models?},
author={Cheng-Ping Hsieh and Simeng Sun and Samuel Kriman and Shantanu Acharya and Dima Rekesh and Fei Jia and Boris Ginsburg},
booktitle={First Conference on Language Modeling},
year={2024},
url={https://openreview.net/forum?id=kIoBbc76Sy}
}

@inproceedings{lcb,
title={LiveCodeBench: Holistic and Contamination Free Evaluation of Large Language Models for Code},
author={Naman Jain and King Han and Alex Gu and Wen-Ding Li and Fanjia Yan and Tianjun Zhang and Sida Wang and Armando Solar-Lezama and Koushik Sen and Ion Stoica},
booktitle={The Thirteenth International Conference on Learning Representations},
year={2025},
url={https://openreview.net/forum?id=chfJJYC3iL}
}

@inproceedings{mmlupro,
 author = {Wang, Yubo and Ma, Xueguang and Zhang, Ge and Ni, Yuansheng and Chandra, Abhranil and Guo, Shiguang and Ren, Weiming and Arulraj, Aaran and He, Xuan and Jiang, Ziyan and Li, Tianle and Ku, Max and Wang, Kai and Zhuang, Alex and Fan, Rongqi and Yue, Xiang and Chen, Wenhu},
 booktitle = {Advances in Neural Information Processing Systems},
 doi = {10.52202/079017-3018},
 editor = {A. Globerson and L. Mackey and D. Belgrave and A. Fan and U. Paquet and J. Tomczak and C. Zhang},
 pages = {95266--95290},
 publisher = {Curran Associates, Inc.},
 title = {MMLU-Pro: A More Robust and Challenging Multi-Task Language Understanding Benchmark},
 url = {https://proceedings.neurips.cc/paper_files/paper/2024/file/ad236edc564f3e3156e1b2feafb99a24-Paper-Datasets_and_Benchmarks_Track.pdf},
 volume = {37},
 year = {2024}
}

@misc{falconh1r,
      title={Falcon-H1R: Pushing the Reasoning Frontiers with a Hybrid Model for Efficient Test-Time Scaling}, 
      author={Falcon LLM Team and Iheb Chaabane and Puneesh Khanna and Suhail Mohmad and Slim Frikha and Shi Hu and Abdalgader Abubaker and Reda Alami and Mikhail Lubinets and Mohamed El Amine Seddik and Hakim Hacid},
      year={2026},
      eprint={2601.02346},
      archivePrefix={arXiv},
      primaryClass={cs.AI},
      url={https://arxiv.org/abs/2601.02346}, 
}

@misc{ministral3,
      title={Ministral 3}, 
      author={Alexander H. Liu and Kartik Khandelwal and Sandeep Subramanian and Victor Jouault and Abhinav Rastogi and Adrien Sadé and Alan Jeffares and Albert Jiang and Alexandre Cahill and Alexandre Gavaudan and others},
      year={2026},
      eprint={2601.08584},
      archivePrefix={arXiv},
      primaryClass={cs.CL},
      url={https://arxiv.org/abs/2601.08584}, 
}

@misc{gptq,
      title={GPTQ: Accurate Post-Training Quantization for Generative Pre-trained Transformers}, 
      author={Elias Frantar and Saleh Ashkboos and Torsten Hoefler and Dan Alistarh},
      year={2023},
      eprint={2210.17323},
      archivePrefix={arXiv},
      primaryClass={cs.LG},
      url={https://arxiv.org/abs/2210.17323}, 
}

@inproceedings{gatedattention,
title={Gated Attention for Large Language Models: Non-linearity, Sparsity, and Attention-Sink-Free},
author={Zihan Qiu and Zekun Wang and Bo Zheng and Zeyu Huang and Kaiyue Wen and Songlin Yang and Rui Men and Le Yu and Fei Huang and Suozhi Huang and Dayiheng Liu and Jingren Zhou and Junyang Lin},
booktitle={The Thirty-ninth Annual Conference on Neural Information Processing Systems},
year={2025},
url={https://openreview.net/forum?id=1b7whO4SfY}
}

@misc{rope,
      title={RoFormer: Enhanced Transformer with Rotary Position Embedding}, 
      author={Jianlin Su and Yu Lu and Shengfeng Pan and Ahmed Murtadha and Bo Wen and Yunfeng Liu},
      year={2023},
      eprint={2104.09864},
      archivePrefix={arXiv},
      primaryClass={cs.CL},
      url={https://arxiv.org/abs/2104.09864}, 
}

@misc{qwen3-next,
    title={{Qwen3-{Next}: Towards Ultimate Training {\&} Inference Efficiency}},
    author={{Qwen Team}},
    url={https://qwen.ai/blog?id=4074cca80393150c248e508aa62983f9cb7d27cd},
    year={2025},
}

@misc{kl-guided-layer-selection,
      title={Distilling to Hybrid Attention Models via KL-Guided Layer Selection}, 
      author={Yanhong Li and Songlin Yang and Shawn Tan and Mayank Mishra and Rameswar Panda and Jiawei Zhou and Yoon Kim},
      year={2025},
      eprint={2512.20569},
      archivePrefix={arXiv},
      primaryClass={cs.CL},
      url={https://arxiv.org/abs/2512.20569}, 
}

@misc{jetnemotron,
      title={Jet-Nemotron: Efficient Language Model with Post Neural Architecture Search}, 
      author={Yuxian Gu and Qinghao Hu and Shang Yang and Haocheng Xi and Junyu Chen and Song Han and Han Cai},
      year={2025},
      eprint={2508.15884},
      archivePrefix={arXiv},
      primaryClass={cs.CL},
      url={https://arxiv.org/abs/2508.15884}, 
}

@misc{rad,
      title={RAD: Redundancy-Aware Distillation for Hybrid Models via Self-Speculative Decoding}, 
      author={Yuichiro Hoshino and Hideyuki Tachibana and Muneyoshi Inahara and Hiroto Takegawa},
      year={2025},
      eprint={2505.22135},
      archivePrefix={arXiv},
      primaryClass={cs.CL},
      url={https://arxiv.org/abs/2505.22135}, 
}

@inproceedings{qknorm,
  title={Query-key normalization for transformers},
  author={Henry, Alex and Dachapally, Prudhvi Raj and Pawar, Shubham Shantaram and Chen, Yuxuan},
  booktitle={Findings of the Association for Computational Linguistics: EMNLP 2020},
  pages={4246--4253},
  year={2020}
}

@inproceedings{mambainllama,
 author = {Wang, Junxiong and Paliotta, Daniele and May, Avner and Rush, Alexander M. and Dao, Tri},
 booktitle = {Advances in Neural Information Processing Systems},
 doi = {10.52202/079017-1996},
 editor = {A. Globerson and L. Mackey and D. Belgrave and A. Fan and U. Paquet and J. Tomczak and C. Zhang},
 pages = {62432--62457},
 publisher = {Curran Associates, Inc.},
 title = {The Mamba in the Llama: Distilling and Accelerating Hybrid Models},
 url = {https://proceedings.neurips.cc/paper_files/paper/2024/file/723933067ad315269b620bc0d2c05cba-Paper-Conference.pdf},
 volume = {37},
 year = {2024}
}

@misc{nemotronnano2,
      title={NVIDIA Nemotron Nano 2: An Accurate and Efficient Hybrid Mamba-Transformer Reasoning Model}, 
      author={{NVIDIA} and Aarti Basant and Abhijit Khairnar and Abhijit Paithankar and Abhinav Khattar and Adithya Renduchintala and Aditya Malte and Akhiad Bercovich and Akshay Hazare and Alejandra Rico and others},
      year={2025},
      eprint={2508.14444},
      archivePrefix={arXiv},
      primaryClass={cs.CL},
      url={https://arxiv.org/abs/2508.14444}, 
}

@misc{falconh1,
      title={Falcon-H1: A Family of Hybrid-Head Language Models Redefining Efficiency and Performance}, 
      author={Jingwei Zuo and Maksim Velikanov and Ilyas Chahed and Younes Belkada and Dhia Eddine Rhayem and Guillaume Kunsch and Hakim Hacid and Hamza Yous and Brahim Farhat and Ibrahim Khadraoui and Mugariya Farooq and Giulia Campesan and Ruxandra Cojocaru and Yasser Djilali and Shi Hu and Iheb Chaabane and Puneesh Khanna and Mohamed El Amine Seddik and Ngoc Dung Huynh and Phuc Le Khac and Leen AlQadi and Billel Mokeddem and Mohamed Chami and Abdalgader Abubaker and Mikhail Lubinets and Kacper Piskorski and Slim Frikha},
      year={2025},
      eprint={2507.22448},
      archivePrefix={arXiv},
      primaryClass={cs.CL},
      url={https://arxiv.org/abs/2507.22448}, 
}

@misc{minicpm4,
      title={MiniCPM4: Ultra-Efficient LLMs on End Devices}, 
      author={MiniCPM-Team and Chaojun Xiao and Yuxuan Li and Xu Han and Yuzhuo Bai and Jie Cai and Haotian Chen and Wentong Chen and Xin Cong and Ganqu Cui and Ning Ding and others},
      year={2025},
      eprint={2506.07900},
      archivePrefix={arXiv},
      primaryClass={cs.CL},
      url={https://arxiv.org/abs/2506.07900}, 
}

@misc{halo,
      title={Hybrid Linear Attention Done Right: Efficient Distillation and Effective Architectures for Extremely Long Contexts}, 
      author={Yingfa Chen and Zhen Leng Thai and Zihan Zhou and Zhu Zhang and Xingyu Shen and Shuo Wang and Chaojun Xiao and Xu Han and Zhiyuan Liu},
      year={2026},
      eprint={2601.22156},
      archivePrefix={arXiv},
      primaryClass={cs.CL},
      url={https://arxiv.org/abs/2601.22156}, 
}

@misc{infllmv2,
      title={InfLLM-V2: Dense-Sparse Switchable Attention for Seamless Short-to-Long Adaptation}, 
      author={Weilin Zhao and Zihan Zhou and Zhou Su and Chaojun Xiao and Yuxuan Li and Yanghao Li and Yudi Zhang and Weilun Zhao and Zhen Li and Yuxiang Huang and Ao Sun and Xu Han and Zhiyuan Liu},
      year={2025},
      eprint={2509.24663},
      archivePrefix={arXiv},
      primaryClass={cs.CL},
      url={https://arxiv.org/abs/2509.24663}, 
}

@inproceedings{infllm,
 author = {Xiao, Chaojun and Zhang, Pengle and Han, Xu and Xiao, Guangxuan and Lin, Yankai and Zhang, Zhengyan and Liu, Zhiyuan and Sun, Maosong},
 booktitle = {Advances in Neural Information Processing Systems},
 doi = {10.52202/079017-3801},
 editor = {A. Globerson and L. Mackey and D. Belgrave and A. Fan and U. Paquet and J. Tomczak and C. Zhang},
 pages = {119638--119661},
 publisher = {Curran Associates, Inc.},
 title = {InfLLM: Training-Free Long-Context Extrapolation for LLMs with an Efficient Context Memory},
 url = {https://proceedings.neurips.cc/paper_files/paper/2024/file/d842425e4bf79ba039352da0f658a906-Paper-Conference.pdf},
 volume = {37},
 year = {2024}
}

@inproceedings{nsa,
    title = "Native Sparse Attention: Hardware-Aligned and Natively Trainable Sparse Attention",
    author = "Yuan, Jingyang  and
      Gao, Huazuo  and
      Dai, Damai  and
      Luo, Junyu  and
      Zhao, Liang  and
      Zhang, Zhengyan  and
      Xie, Zhenda  and
      Wei, Yuxing  and
      Wang, Lean  and
      Xiao, Zhiping  and
      Wang, Yuqing  and
      Ruan, Chong  and
      Zhang, Ming  and
      Liang, Wenfeng  and
      Zeng, Wangding",
    booktitle = "Proceedings of the 63rd Annual Meeting of the Association for Computational Linguistics (Volume 1: Long Papers)",
    year = "2025",
    url = "https://aclanthology.org/2025.acl-long.1126/",
}

@inproceedings{gateddeltanet,
title={Gated Delta Networks: Improving Mamba2 with Delta Rule},
author={Songlin Yang and Jan Kautz and Ali Hatamizadeh},
booktitle={The Thirteenth International Conference on Learning Representations},
year={2025},
url={https://openreview.net/forum?id=r8H7xhYPwz}
}

@inproceedings{deltanet,
title={Parallelizing Linear Transformers with the Delta Rule over Sequence Length},
author={Songlin Yang and Bailin Wang and Yu Zhang and Yikang Shen and Yoon Kim},
booktitle={The Thirty-eighth Annual Conference on Neural Information Processing Systems},
year={2024},
url={https://openreview.net/forum?id=y8Rm4VNRPH}
}

@inproceedings{rwkv,
    title = "{RWKV}: Reinventing {RNN}s for the Transformer Era",
    author = "Peng, Bo  and
      Alcaide, Eric  and
      Anthony, Quentin  and
      Albalak, Alon  and
      Arcadinho, Samuel  and
      Biderman, Stella  and
      Cao, Huanqi  and
      Cheng, Xin  and
      Chung, Michael  and
      Derczynski, Leon  and
      Du, Xingjian  and
      Grella, Matteo  and
      Gv, Kranthi  and
      He, Xuzheng  and
      Hou, Haowen  and
      Kazienko, Przemyslaw  and
      Kocon, Jan  and
      Kong, Jiaming  and
      Koptyra, Bart{\l}omiej  and
      Lau, Hayden  and
      Lin, Jiaju  and
      Mantri, Krishna Sri Ipsit  and
      Mom, Ferdinand  and
      Saito, Atsushi  and
      Song, Guangyu  and
      Tang, Xiangru  and
      Wind, Johan  and
      Wo{\'z}niak, Stanis{\l}aw  and
      Zhang, Zhenyuan  and
      Zhou, Qinghua  and
      Zhu, Jian  and
      Zhu, Rui-Jie",
    booktitle = "Findings of the Association for Computational Linguistics: EMNLP 2023",
    year = "2023",
    url = "https://aclanthology.org/2023.findings-emnlp.936/",
}

@inproceedings{mamba,
title={Mamba: Linear-Time Sequence Modeling with Selective State Spaces},
author={Albert Gu and Tri Dao},
booktitle={First Conference on Language Modeling},
year={2024},
url={https://openreview.net/forum?id=tEYskw1VY2}
}

@inproceedings{gla,
title={Gated Linear Attention Transformers with Hardware-Efficient Training},
author={Songlin Yang and Bailin Wang and Yikang Shen and Rameswar Panda and Yoon Kim},
booktitle={Forty-first International Conference on Machine Learning},
year={2024},
url={https://openreview.net/forum?id=ia5XvxFUJT}
}

@inproceedings{lightningattention,
title={Various Lengths, Constant Speed: Efficient Language Modeling with Lightning Attention},
author={Zhen Qin and Weigao Sun and Dong Li and Xuyang Shen and Weixuan Sun and Yiran Zhong},
booktitle={Forty-first International Conference on Machine Learning},
year={2024},
url={https://openreview.net/forum?id=Lwm6TiUP4X}
}

@inproceedings{gqa,
    title = "{GQA}: Training Generalized Multi-Query Transformer Models from Multi-Head Checkpoints",
    author = "Ainslie, Joshua  and
      Lee-Thorp, James  and
      de Jong, Michiel  and
      Zemlyanskiy, Yury  and
      Lebron, Federico  and
      Sanghai, Sumit",
    booktitle = "Proceedings of the 2023 Conference on Empirical Methods in Natural Language Processing",
    year = "2023",
    url = "https://aclanthology.org/2023.emnlp-main.298/",
}

@inproceedings{Transformer,
 author = {Vaswani, Ashish and Shazeer, Noam and Parmar, Niki and Uszkoreit, Jakob and Jones, Llion and Gomez, Aidan N and Kaiser, \L ukasz and Polosukhin, Illia},
 booktitle = {Advances in Neural Information Processing Systems},
 editor = {I. Guyon and U. Von Luxburg and S. Bengio and H. Wallach and R. Fergus and S. Vishwanathan and R. Garnett},
 pages = {},
 publisher = {Curran Associates, Inc.},
 title = {Attention is All you Need},
 url = {https://proceedings.neurips.cc/paper_files/paper/2017/file/3f5ee243547dee91fbd053c1c4a845aa-Paper.pdf},
 volume = {30},
 year = {2017}
}

@misc{nemotron3nano,
      title={Nemotron 3 Nano: Open, Efficient Mixture-of-Experts Hybrid Mamba-Transformer Model for Agentic Reasoning}, 
      author={{NVIDIA} and Aaron Blakeman and Aaron Grattafiori and Aarti Basant and Abhibha Gupta and Abhinav Khattar and Adi Renduchintala and Aditya Vavre and Akanksha Shukla and Akhiad Bercovich and Aleksander Ficek and others},
      year={2025},
      eprint={2512.20848},
      archivePrefix={arXiv},
      primaryClass={cs.CL},
      url={https://arxiv.org/abs/2512.20848}, 
}

@misc{kimilinear,
      title={Kimi Linear: An Expressive, Efficient Attention Architecture}, 
      author={{Kimi Team} and Yu Zhang and Zongyu Lin and Xingcheng Yao and Jiaxi Hu and Fanqing Meng and Chengyin Liu and Xin Men and Songlin Yang and Zhiyuan Li and Wentao Li and others},
      year={2025},
      eprint={2510.26692},
      archivePrefix={arXiv},
      primaryClass={cs.CL},
      url={https://arxiv.org/abs/2510.26692}, 
}

@inproceedings{chatdev,
    title = "{C}hat{D}ev: Communicative Agents for Software Development",
    author = "Qian, Chen  and
      Liu, Wei  and
      Liu, Hongzhang  and
      Chen, Nuo  and
      Dang, Yufan  and
      Li, Jiahao  and
      Yang, Cheng  and
      Chen, Weize  and
      Su, Yusheng  and
      Cong, Xin  and
      Xu, Juyuan  and
      Li, Dahai  and
      Liu, Zhiyuan  and
      Sun, Maosong",
    booktitle = "Proceedings of the 62nd Annual Meeting of the Association for Computational Linguistics (Volume 1: Long Papers)",
    year = "2024",
    url = "https://aclanthology.org/2024.acl-long.810/",
}

@misc{agencybench,
      title={AgencyBench: Benchmarking the Frontiers of Autonomous Agents in 1M-Token Real-World Contexts}, 
      author={Keyu Li and Junhao Shi and Yang Xiao and Mohan Jiang and Jie Sun and Yunze Wu and Shijie Xia and Xiaojie Cai and Tianze Xu and Weiye Si and Wenjie Li and Dequan Wang and Pengfei Liu},
      year={2026},
      eprint={2601.11044},
      archivePrefix={arXiv},
      primaryClass={cs.AI},
      url={https://arxiv.org/abs/2601.11044}, 
}

@misc{gaia,
      title={GAIA: a benchmark for General AI Assistants}, 
      author={Grégoire Mialon and Clémentine Fourrier and Craig Swift and Thomas Wolf and Yann LeCun and Thomas Scialom},
      year={2023},
      eprint={2311.12983},
      archivePrefix={arXiv},
      primaryClass={cs.CL},
      url={https://arxiv.org/abs/2311.12983}, 
}

@inproceedings{repobench,
title={RepoBench: Benchmarking Repository-Level Code Auto-Completion Systems},
author={Tianyang Liu and Canwen Xu and Julian McAuley},
booktitle={The Twelfth International Conference on Learning Representations},
year={2024},
url={https://openreview.net/forum?id=pPjZIOuQuF}
}

@inproceedings{swebench,
title={{SWE}-bench: Can Language Models Resolve Real-world Github Issues?},
author={Carlos E Jimenez and John Yang and Alexander Wettig and Shunyu Yao and Kexin Pei and Ofir Press and Karthik R Narasimhan},
booktitle={The Twelfth International Conference on Learning Representations},
year={2024},
url={https://openreview.net/forum?id=VTF8yNQM66}
}

@misc{deepseekcoder,
      title={DeepSeek-Coder: When the Large Language Model Meets Programming -- The Rise of Code Intelligence}, 
      author={Daya Guo and Qihao Zhu and Dejian Yang and Zhenda Xie and Kai Dong and Wentao Zhang and Guanting Chen and Xiao Bi and Y. Wu and Y. K. Li and Fuli Luo and Yingfei Xiong and Wenfeng Liang},
      year={2024},
      eprint={2401.14196},
      archivePrefix={arXiv},
      primaryClass={cs.SE},
      url={https://arxiv.org/abs/2401.14196}, 
}

@inproceedings{llmxmapreduce,
    title = "{LLM}$\times${M}ap{R}educe: Simplified Long-Sequence Processing using Large Language Models",
    author = "Zhou, Zihan  and
      Li, Chong  and
      Chen, Xinyi  and
      Wang, Shuo  and
      Chao, Yu  and
      Li, Zhili  and
      Wang, Haoyu  and
      Shi, Qi  and
      Tan, Zhixing  and
      Han, Xu  and
      Shi, Xiaodong  and
      Liu, Zhiyuan  and
      Sun, Maosong",
    booktitle = "Proceedings of the 63rd Annual Meeting of the Association for Computational Linguistics (Volume 1: Long Papers)",
    year = "2025",
    url = "https://aclanthology.org/2025.acl-long.1341/",
}

@inproceedings{storm,
    title = "Assisting in Writing {W}ikipedia-like Articles From Scratch with Large Language Models",
    author = "Shao, Yijia  and
      Jiang, Yucheng  and
      Kanell, Theodore  and
      Xu, Peter  and
      Khattab, Omar  and
      Lam, Monica",
    booktitle = "Proceedings of the 2024 Conference of the North American Chapter of the Association for Computational Linguistics: Human Language Technologies (Volume 1: Long Papers)",
    year = "2024",
    url = "https://aclanthology.org/2024.naacl-long.347/",
}

@inproceedings{longwriter,
title={LongWriter: Unleashing 10,000+ Word Generation from Long Context {LLM}s},
author={Yushi Bai and Jiajie Zhang and Xin Lv and Linzhi Zheng and Siqi Zhu and Lei Hou and Yuxiao Dong and Jie Tang and Juanzi Li},
booktitle={The Thirteenth International Conference on Learning Representations},
year={2025},
url={https://openreview.net/forum?id=kQ5s9Yh0WI}
}

@inproceedings{longalign,
    title = "{L}ong{A}lign: A Recipe for Long Context Alignment of Large Language Models",
    author = "Bai, Yushi  and
      Lv, Xin  and
      Zhang, Jiajie  and
      He, Yuze  and
      Qi, Ji  and
      Hou, Lei  and
      Tang, Jie  and
      Dong, Yuxiao  and
      Li, Juanzi",
    booktitle = "Findings of the Association for Computational Linguistics: EMNLP 2024",
    year = "2024",
    url = "https://aclanthology.org/2024.findings-emnlp.74/",
}

@inproceedings{gpt3,
 author = {Brown, Tom and Mann, Benjamin and Ryder, Nick and Subbiah, Melanie and Kaplan, Jared D and Dhariwal, Prafulla and Neelakantan, Arvind and Shyam, Pranav and Sastry, Girish and Askell, Amanda and others},
 booktitle = {Advances in Neural Information Processing Systems},
 editor = {H. Larochelle and M. Ranzato and R. Hadsell and M.F. Balcan and H. Lin},
 pages = {1877--1901},
 publisher = {Curran Associates, Inc.},
 title = {Language Models are Few-Shot Learners},
 url = {https://proceedings.neurips.cc/paper_files/paper/2020/file/1457c0d6bfcb4967418bfb8ac142f64a-Paper.pdf},
 volume = {33},
 year = {2020}
}

@misc{gpt4,
      title={GPT-4 Technical Report}, 
      author={OpenAI and Josh Achiam and Steven Adler and Sandhini Agarwal and Lama Ahmad and Ilge Akkaya and Florencia Leoni Aleman and Diogo Almeida and Janko Altenschmidt and Sam Altman and Shyamal Anadkat and others},
      year={2024},
      eprint={2303.08774},
      archivePrefix={arXiv},
      primaryClass={cs.CL},
      url={https://arxiv.org/abs/2303.08774}, 
}

@misc{gemini25,
      title={Gemini 2.5: Pushing the Frontier with Advanced Reasoning, Multimodality, Long Context, and Next Generation Agentic Capabilities}, 
      author={Gheorghe Comanici and Eric Bieber and Mike Schaekermann and Ice Pasupat and Noveen Sachdeva and Inderjit Dhillon and Marcel Blistein and Ori Ram and Dan Zhang and Evan Rosen and others},
      year={2025},
      eprint={2507.06261},
      archivePrefix={arXiv},
      primaryClass={cs.CL},
      url={https://arxiv.org/abs/2507.06261}, 
}

@misc{qwen3,
      title={Qwen3 Technical Report}, 
      author={An Yang and Anfeng Li and Baosong Yang and Beichen Zhang and Binyuan Hui and Bo Zheng and Bowen Yu and Chang Gao and Chengen Huang and Chenxu Lv and others},
      year={2025},
      eprint={2505.09388},
      archivePrefix={arXiv},
      primaryClass={cs.CL},
      url={https://arxiv.org/abs/2505.09388}, 
}

@misc{llama3,
      title={The Llama 3 Herd of Models}, 
      author={Aaron Grattafiori and Abhimanyu Dubey and Abhinav Jauhri and Abhinav Pandey and Abhishek Kadian and Ahmad Al-Dahle and Aiesha Letman and Akhil Mathur and Alan Schelten and Alex Vaughan and others},
      year={2024},
      eprint={2407.21783},
      archivePrefix={arXiv},
      primaryClass={cs.AI},
      url={https://arxiv.org/abs/2407.21783}, 
}

@misc{deepseekv32,
      title={DeepSeek-V3.2: Pushing the Frontier of Open Large Language Models}, 
      author={DeepSeek-AI and Aixin Liu and Aoxue Mei and Bangcai Lin and Bing Xue and Bingxuan Wang and Bingzheng Xu and Bochao Wu and Bowei Zhang and Chaofan Lin and Chen Dong and others},
      year={2025},
      eprint={2512.02556},
      archivePrefix={arXiv},
      primaryClass={cs.CL},
      url={https://arxiv.org/abs/2512.02556}, 
}

@misc{cmmlu,
      title={CMMLU: Measuring massive multitask language understanding in Chinese}, 
      author={Haonan Li and Yixuan Zhang and Fajri Koto and Yifei Yang and Hai Zhao and Yeyun Gong and Nan Duan and Timothy Baldwin},
      year={2023},
      eprint={2306.09212},
      archivePrefix={arXiv},
      primaryClass={cs.CL}
}

@article{bbh,
  title={Challenging big-bench tasks and whether chain-of-thought can solve them},
  author={Suzgun, Mirac and Scales, Nathan and Sch{\"a}rli, Nathanael and Gehrmann, Sebastian and Tay, Yi and Chung, Hyung Won and Chowdhery, Aakanksha and Le, Quoc V and Chi, Ed H and Zhou, Denny and others},
  journal={arXiv preprint arXiv:2210.09261},
  year={2022}
}

@article{math,
  title={Measuring mathematical problem solving with the math dataset},
  author={Hendrycks, Dan and Burns, Collin and Kadavath, Saurav and Arora, Akul and Basart, Steven and Tang, Eric and Song, Dawn and Steinhardt, Jacob},
  journal={arXiv preprint arXiv:2103.03874},
  year={2021}
}

@article{mbpp,
  title={Program synthesis with large language models},
  author={Austin, Jacob and Odena, Augustus and Nye, Maxwell and Bosma, Maarten and Michalewski, Henryk and Dohan, David and Jiang, Ellen and Cai, Carrie and Terry, Michael and Le, Quoc and others},
  journal={arXiv preprint arXiv:2108.07732},
  year={2021}
}

@article{humaneval,
  title={Evaluating large language models trained on code},
  author={Chen, Mark and Tworek, Jerry and Jun, Heewoo and Yuan, Qiming and Pinto, Henrique Ponde De Oliveira and Kaplan, Jared and Edwards, Harri and Burda, Yuri and Joseph, Nicholas and Brockman, Greg and others},
  journal={arXiv preprint arXiv:2107.03374},
  year={2021}
}

@misc{opencompass,
    title={OpenCompass: A Universal Evaluation Platform for Foundation Models},
    author={OpenCompass Contributors},
    howpublished = {\url{https://github.com/open-compass/opencompass}},
    year={2023}
}

@misc{wang2026datasciencetechnologyagi,
  title={Data Science and Technology Towards AGI Part I: Tiered Data Management},
  author={Yudong Wang and Zixuan Fu and Hengyu Zhao and Chen Zhao and Chuyue Zhou and Xinle Lin and Hongya Lyu and Shuaikang Xue and Yi Yi and Yingjiao Wang and Zhi Zheng and Yuzhou Zhang and Jie Zhou and Chaojun Xiao and Xu Han and Zhiyuan Liu and Maosong Sun},
  year={2026},
  eprint={2602.09003},
  archivePrefix={arXiv},
  primaryClass={cs.AI},
  url={https://arxiv.org/abs/2602.09003}
}
